\documentclass[letterpaper]{article} 
\usepackage{aaai25}  
\usepackage{times}  
\usepackage{helvet}  
\usepackage{courier}  
\usepackage[hyphens]{url}  
\usepackage{graphicx} 
\usepackage{natbib}  
\usepackage{caption} 
\usepackage{algorithm}
\usepackage{algorithmic}
\usepackage{amsmath, amsfonts, amssymb}
\usepackage{tabularray}

\usepackage{newfloat}
\usepackage{listings}
\DeclareCaptionStyle{ruled}{labelfont=normalfont,labelsep=colon,strut=off} 
\floatstyle{ruled}
\newfloat{listing}{tb}{lst}{}
\floatname{listing}{Listing}
\title{$p$SVM: Soft-margin SVMs with $p$-norm Hinge Loss}
\author {
    Haoxiang Sun
}
\affiliations{
    School of Information, Renmin Univerity of China, Beijing, China\\
    hxiang.sun@gmail.com
}

\begin{document}

\maketitle

\begin{abstract}
Support Vector Machines (SVMs) based on hinge loss have been extensively discussed and applied to various binary classification tasks. These SVMs achieve a balance between margin maximization and the minimization of slack due to outliers. Although many efforts have been dedicated to enhancing the performance of SVMs with hinge loss, studies on $p$SVMs, soft-margin SVMs with $p$-norm hinge loss, remain relatively scarce. In this paper, we explore the properties, performance, and training algorithms of $p$SVMs. We first derive the generalization bound of $p$SVMs, then formulate the dual optimization problem, comparing it with the traditional approach. Furthermore, we discuss a generalized version of the Sequential Minimal Optimization (SMO) algorithm, $p$SMO, to train our $p$SVM model. Comparative experiments on various datasets, including binary and multi-class classification tasks, demonstrate the effectiveness and advantages of our $p$SVM model and the $p$SMO method. Code is available at https://github.com/CoderBak/pSVM.
\end{abstract}

%

\section{Introduction}

Support Vector Machine (SVM) algorithm \cite{boser1992training}, which is known for its simplicity and effectiveness, has been extensively utilized not only in practical classification tasks, such as image classification \cite{cortes1995support}, healthcare \cite{braga2019automatic}, and cybersecurity \cite{krishnaveni2020anomaly}, but also in learning algorithms, such as federated learning \cite{wang2024turbosvm}, and deep stacking networks \cite{wang2019svm}.

Originally designed for binary classification, SVMs can be extended to handle multiclass classification tasks using One versus One (OvO) and One versus Rest (OvR) techniques. In a $k$-class classification scenario, OvO constructs $k(k-1)/2$ base classifiers to determine hyperplanes separating each pair of classes, whereas OvR builds $k$ base classifiers to distinguish each class from the rest. Previous studies have indicated that OvO generally outperforms OvR in large-scale problems \cite{hsu2002comparison}. However, the efficiency of OvO is constrained by the training time of base classifiers, and its performance heavily depends on the accuracy of these classifiers. Therefore, improving training efficiency and accuracy in SVMs can enhance the effectiveness of multiclass classification.

In the context of soft-margin SVMs \cite{cortes1995support}, the parameter $C\ge 0$ determines the trade-off between margin-maximization and the minimization of slack penalty \cite{mohri2018foundations}. While most studies focus on SVMs with \textit{hinge loss}, defined as $C\sum_{i=1}^m \xi_i$, recent studies have also explored SVMs with \textit{quadratic hinge loss}. For instance, \citealp{lee2013study} extended Crammer and Singer's multiclass SVM \cite{crammer2001algorithmic} to apply L2 loss, achieving superior accuracy but at the cost of longer training time. This suggests the potential of introducing a more generalized $p$-norm hinge loss $C\sum_{i=1}^m \xi_i^p$ to improve the accuracy and efficiency of SVMs. Nonetheless, studies on $p$-norm hinge loss in SVMs remain limited.

To address these challenges, we propose a novel model, $p$SVM, i.e. soft-margin SVMs with $p$-norm hinge loss. This model generalizes both L1 and L2 loss SVMs and allows the selection of an optimal parameter $p$ to enhance the performance. However, this extension may increase training time. To mitigate this problem, we introduce the $p$SMO method to efficiently solve the optimization problem associated with $p$SVM training. Our approach facilitates the development of effective binary and multiclass classifiers.

Our contributions are summarized as follows.

\begin{itemize}

\item We introduce a novel model $p$SVM, by incorporating a hyperparameter $p\ge 1$ into soft-margin SVMs, which enhances the flexibility in balancing margin maximization with the minimization of slack caused by outliers. Furthermore, we establish a generalization bound through the lens of margin theory, offering theoretical insight into the motivation behind the $p$-norm hinge loss.

\item We develop the $p$SMO method to address the challenges of solving the dual optimization problem in the $p$SVM model. In particular, we present practical implementations of $1.5$SMO and $2$SMO, and integrate them with the one-vs-one (OvO) strategy for effective multiclass classification.

\item Through extensive experiments on diverse datasets, our proposed methods demonstrate significant improvements in classification performance. These experiments on binary and multiclass classification tasks highlight the properties of $p$SVM models and the effectiveness of $p$SMO methods.

\end{itemize}

This paper is organized as follows. In Section 2, we review the fundamental concepts relevant to our topic, define key terms, and introduce our proposed $p$SVM model. Section 3 explores the generalization bound and dual optimization problem of the $p$SVM model, proving that it minimizes the upper-bound of the generalization error. In Section 4, we present the $p$SMO method, aiming at effectively solving the dual optimization problem of $p$SVMs, and we described how multiclass classifiers can be constructed based on the $p$SVM model and $p$SMO method. Section 5 presents experimental results, highlighting the properties of the $p$SVM model in binary classifications and evaluating the efficiency of the $p$SMO method in comparison to state-of-the-art algorithms on multiclass classification tasks. Section 6 then concludes the paper. 

\section{Preliminaries}

Before delving into the main discussion, we first review some fundamental concepts relevant to our topic. The set $\{1,2,\cdots, n\}$ is abbreviated as $[n]$ for simplicity.

\subsection{SVMs for binary classification}

A classification task can be framed as an optimization problem: Given a training sample $S$, the learner aims to determine a hypothesis, or classifier, $h\in\mathcal{H}$, that minimizes the generalization error. In the simplest case involving hyperplane classifiers, the objective is to find a hyperplane that optimally separates data points belonging to different classes, which is precisely the task of the Support Vector Machine (SVM) algorithm \cite{boser1992training}.

For training samples that are not linearly separable, the soft-margin SVM \cite{cortes1995support} was introduced, balancing the trade-off between maximizing the margin and minimizing the slack variables associated with outliers. This optimization problem is formulated as:
\begin{equation}\label{eq1}
\begin{split}
    \min_{\mathbf{w}, b}&\  \frac{1}{2}\Vert \mathbf{w}\Vert^2 + C\sum_{i=1}^m \xi_i,\\
    \text{s.t.} &\  y_i(\mathbf{w}\cdot \mathbf{x}_i + b)\ge 1-\xi_i, i\in [n],
\end{split}
\end{equation}
where slack variables can also be viewed as hinge loss, i.e.
\begin{equation}\label{hingeopt}
    \min_{\mathbf{w}, b}\  \frac{1}{2}\Vert \mathbf{w}\Vert^2 + C\sum_{i=1}^m \max\{0, 1- y_i(\mathbf{w}\cdot \mathbf{x}_i + b)\}.
\end{equation}

By applying the Karush–Kuhn–Tucker (KKT) conditions and kernel methods \cite{boser1992training} to Eq. (\ref{eq1}), we obtain the dual optimization problem \cite{mohri2018foundations},
\begin{equation}\label{eq2}
\begin{split}
    \max_{\boldsymbol{\alpha}}&\  \sum_{i=1}^m \alpha_i - \dfrac{1}{2} \sum_{i,j=1}^m \alpha_i \alpha_j y_i y_j K(\mathbf{x}_i,\mathbf{x}_j),\\
    \text{s.t.} &\ 0\le \alpha_i \le C \wedge \sum_{i=1}^m \alpha_i y_i = 0, i\in [n],
\end{split}
\end{equation}
and its solution directly yields the classifier:
\begin{equation}\label{eq3}
h(\mathbf{x}) = \mathrm{sgn} \ \varphi(\mathbf{x}) = \mathrm{sgn}\left(\sum_{i=1}^m \alpha_i y_i K(\mathbf{x}_i, \mathbf{x}) + b\right).
\end{equation}

In the context of SVMs, the \textit{confidence margin} for each data point $x$ in the training set is defined as its distance from the hyperplane $h$. The SVM algorithm aims to identify a hyperplane that maximizes the confidence margin for the least confident data point. Consequently, the behaviour of soft-margin SVM can be estimated through margin theory.

\subsection{Margin Theory}

The following definitions extend the concept of \textit{confidence margin} by introducing a hyperparameter $p\ge1$.

\noindent\textbf{Definition 1} (Margin loss function) \textit{For any $\rho >0$, the $\rho$-margin loss under $p$-norm is the function $L_{p,\rho}:\mathbb{R}\times \mathbb{R}\to \mathbb{R}_{+}$ defined for all $y,y'\in \mathbb{R}$ by $L_{p,\rho}=\Phi_{p,\rho}(yy')$ with}
\begin{equation}\label{eq4}
\begin{split}
\Phi_{p,\rho}(x)&=\min\left(1,\max\left(0,\left(1-\frac{x}{\rho}\right)^p\right)\right)\\
&=\begin{cases}
1, & \text{if } x\le 0\\
(1-x/\rho)^p, & \text{if } 0\le x \le \rho\\
0, & \text{if } x \ge \rho.
\end{cases}
\end{split}
\end{equation}

\noindent\textbf{Definition 2} (Empirical margin loss) \textit{Given a sample $S=(x_1, \cdots, x_m)$ and a hypothesis $h$, the empirical margin loss is defined by}
\begin{equation}\label{eq5}
\widehat{R}_{S,p,\rho}(h) = \dfrac{1}{m} \sum_{i=1}^m \Phi_{p, \rho}(y_ih(x_i)).
\end{equation}

\noindent\textbf{Definition 3} (Generalization error) \textit{The generalization error of a hypothesis $h$ is defined by}
\begin{equation}\label{eq7}
R_{p,\rho}(h) = \underset{S\sim \mathcal{D}^m}{\mathbb{E}}\widehat{R}_{S,p,\rho}(h).
\end{equation}

\begin{figure}[t]
\centering
\includegraphics[width=0.9\columnwidth]{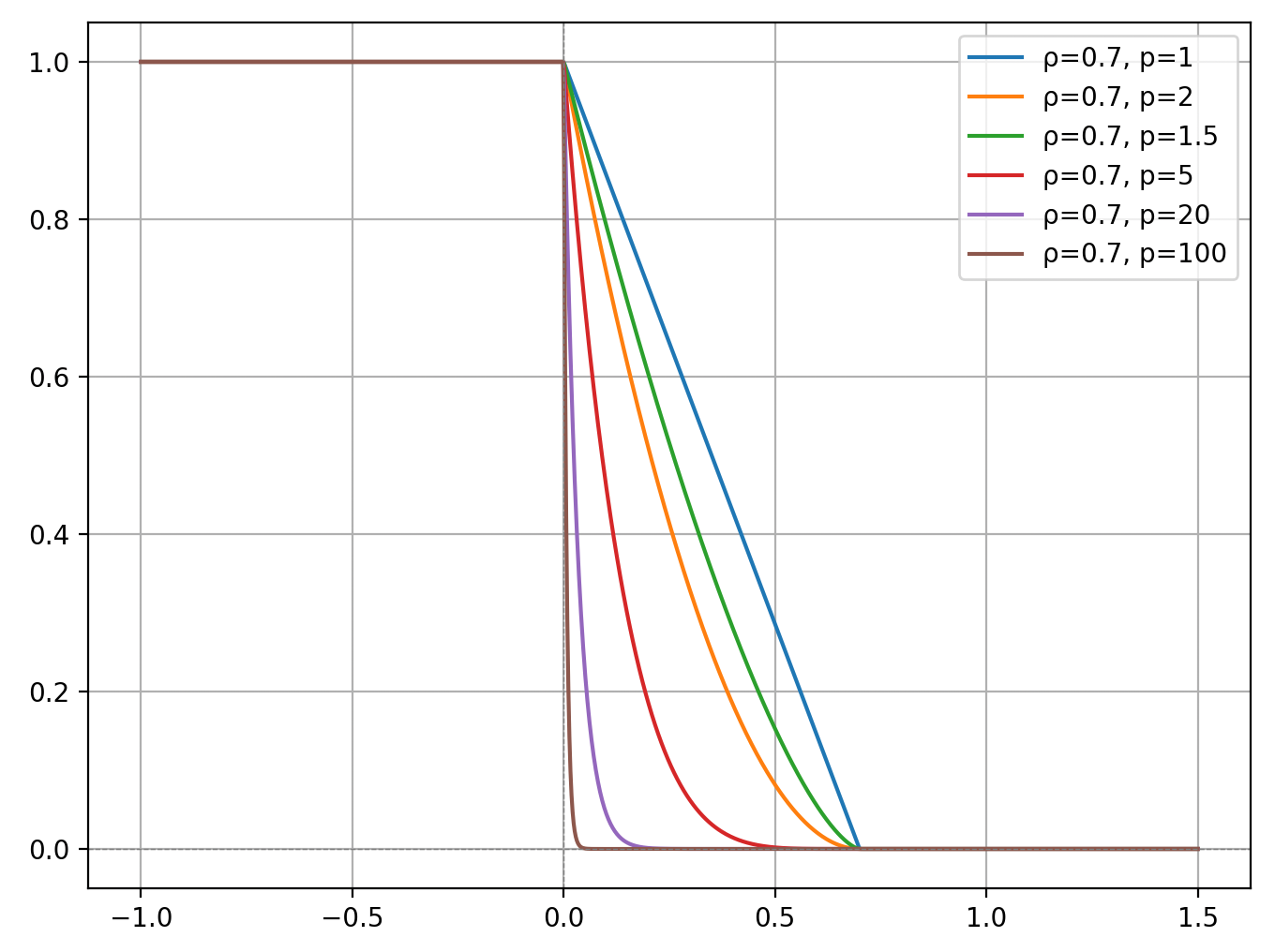} 
\caption{$\Phi_{p,\rho}$ under different $p$-values.}
\label{figloss}
\end{figure}

Fig. \ref{figloss} provides an intuitive view of the loss function. Note that $\Phi_{p,\rho}$ is upper-bounded by the $p$-norm hinge loss:
\begin{equation}\label{eq6}
\Phi_{p,\rho}(x) \le \max\left( 0, \left(1-\dfrac{x}{\rho}\right)^p\right).
\end{equation}

\subsection{$p$SVM Model}

Our $p$SVM model, which generalizes the soft-margin SVM by introducing the $p$-norm hinge loss, can be formulated as the following optimization problem, where $p\ge 1$ is a hyperparameter that can be selected via cross-validation.
\begin{equation}\label{eq1*}
\begin{split}
    \min_{\mathbf{w}, b}&\  \frac{1}{2}\Vert \mathbf{w}\Vert^2 + C\sum_{i=1}^m \xi_i^p,\\
    \text{s.t.} &\  y_i(\mathbf{w}\cdot \mathbf{x}_i + b)\ge 1-\xi_i, i\in [n],
\end{split}
\end{equation}
thus, our $p$SVM model is actually the soft-margin SVM with $p$-norm hinge loss ($p\ge 1$), i.e.
\begin{equation}\label{hingeopt*}
    \min_{\mathbf{w}, b}\  \frac{1}{2}\Vert \mathbf{w}\Vert^2 + C\sum_{i=1}^m \max(0, 1- y_i(\mathbf{w}\cdot \mathbf{x}_i + b))^p,
\end{equation}
which can be used to train binary classifiers.

\section{Theoretical Derivation}

In this section, we discuss the generalization bound and dual optimization problem of our $p$SVM model.

\subsection{Generalization Bound}

\citealp{mohri2018foundations} derived the following bounds on the generalization error and Rademacher complexity \cite{bartlett2002rademacher}.

\noindent\textbf{Theorem 1} \textit{Let $\mathcal{H}$ be a set of real-valued functions. Fix $\rho > 0$, then, for any $\delta>0$, with probability at least $1-\delta$, the following holds for all $h\in\mathcal{H}$}:
\begin{equation}\label{eqx}
R_{1,\rho}(h)\le\widehat{R}_{S,1,\rho}(h) + \dfrac{2}{\rho}\mathfrak{R}_m(\mathcal{H}) + \sqrt{\dfrac{\log \frac{2}{\delta}}{2m}}.
\end{equation}

\noindent\textbf{Theorem 2} \textit{Let $S\subseteq\{\mathbf{x}: \Vert\mathbf{x}\Vert\le r\}$ be a sample of size $m$ and let $\mathcal{H}=\{\mathbf{x}\mapsto \mathbf{w}\cdot \mathbf{x}: \Vert\mathbf{w}\Vert \le \Lambda\}$. Then, the empirical Rademacher complexity of $\mathcal{H}$ can be bounded as follows}:
\begin{equation}\label{eqy}
    \widehat{\mathfrak{R}}_{S}(\mathcal{H})\le \sqrt{\dfrac{r^2\Lambda^2}{m}}.
\end{equation}

However, these results pertain specifically to the tradition soft-margin SVM, where $p=1$. In our work, we extend these findings by deriving new theoretical results on the generalization bounds for $p$SVMs.

\noindent\textbf{Lemma 1} \textit{$\Phi_{p,\rho}$ is $p/\rho$-lipschitz}.

This lemma, which highlights an important property of the margin loss function, serves as a foundation for the following theorem.

\noindent\textbf{Theorem 3} \textit{Let $\mathcal{H} = \{\mathbf{x}\mapsto \mathbf{w}\cdot\mathbf{x} + b: \Vert\mathbf{w}\Vert\le 1\}$ and assume that $\mathcal{X} \subseteq \{\mathbf{x}:\Vert\mathbf{x}\Vert\le r\}$. Fix $r$, then, for any $\delta > 0$, with probability at least $1-\delta$ over the choice of a sample $S$ of size $m$, the following holds for any $h\in \mathcal{H}$ and $\rho\in (0,r]$}:
\begin{equation}\label{bound1}
\begin{split}
R_{p,\rho}(h)\le &\ \widehat{R}_{S,p,\rho}(h)
		\\ &+ {4p}\sqrt{\dfrac{r^2/\rho^2}{m}}+\sqrt{\dfrac{\log \log_2 \frac{2r}{\rho}}{m}} + \sqrt{\dfrac{\log \frac{2}{\delta}}{2m}}.
\end{split}
\end{equation}

Theorem 3 leads to the following corollary, which establishes the generalization bound of $p$SVMs.

\noindent\textbf{Corollary 1} (Generalization bound of $p$SVMs) \textit{Suppose $\mathcal{H} = \{\mathbf{x}\mapsto \mathbf{w}\cdot\mathbf{x} + b: \Vert\mathbf{w}\Vert\le 1/\rho\}$ and assume that $\mathcal{X} \subseteq \{\mathbf{x}:\Vert\mathbf{x}\Vert\le r\}$. Fix $r>0$, then, for any $\delta > 0$, with probability at least $1-\delta$ over the choice of a sample $S$ of size $m$, the following holds for any $h\in \mathcal{H}$ and $\rho\in(0,r]$}:
\begin{equation}\label{bound}
\begin{split}
R_{p,\rho}(h)\le &\ \dfrac{1}{m}\sum_{i=1}^m (\max(0, 1-y_i(\mathbf{w}\cdot \mathbf{x}_i + b)))^p
		\\ &+ 4p\sqrt{\dfrac{r^2/\rho^2}{m}}+\sqrt{\dfrac{\log \log_2 \frac{2r}{\rho}}{m}} + \sqrt{\dfrac{\log \frac{2}{\delta}}{2m}}.
\end{split}
\end{equation}

Following a similar method in \citealp{mohri2018foundations}, we analyze our result as follows. Consider the right-hand side of Eq. (\ref{bound}). As $p\ge 1$ increases, the first term decreases while the second term increases. This trade-off highlights the significance of the hyperparameter $p$ in our $p$SVM model. By appropriately selecting $p$, one can achieve an optimal balance between these two terms, thus minimizing the right-hand side of Eq. (\ref{bound}).

Since only the first term of the right-hand side depends on $\mathbf{w}$, for any $\rho > 0$, one can choose the best hyperplane by selecting $\mathbf{w}$ as the solution of the following optimization problem:
\begin{equation}\label{bound2}
\begin{split}
\min_{\Vert\mathbf{w}\Vert^2 \le \frac{1}{\rho^2}} \dfrac{1}{m} \sum_{i=1}^m \max(0, 1-y_i(\mathbf{w}\cdot \mathbf{x}_i + b))^p.
\end{split}
\end{equation}

Introducing a Lagrange variable $\lambda \ge 0$, the optimization problem can be equivalently written as
\begin{equation}\label{bound3}
\begin{split}
\min_{\mathbf{w}} \lambda \Vert\mathbf{w}\Vert^2 + \dfrac{1}{m} \sum_{i=1}^m \max(0, 1-y_i(\mathbf{w}\cdot \mathbf{x}_i + b))^p,
\end{split}
\end{equation}
which precisely coincides with our $p$SVM model Eq. (\ref{hingeopt*}), indicating that the learning performance of $p$SVMs is guaranteed by minimizing the right-hand side of Eq. (\ref{bound}).

\subsection{Dual Optimization Problem}

Applying the Karush–Kuhn–Tucker (KKT) conditions and kernel methods to Eq. (\ref{eq1*}) yields the dual optimization problem (which is concave) as follows:
\begin{equation}\label{dual}
\begin{split}
    \max_{\boldsymbol{\alpha}}\ \ \ &\sum_{i=1}^m \alpha_i - \theta\sum_{i=1}^m \alpha_i^\gamma - \dfrac{1}{2} \sum_{i,j=1}^m \alpha_i \alpha_j y_i y_j K(\mathbf{x}_i, \mathbf{x}_j), \\
    \text{s.t.} &\ \alpha_i \ge 0 \wedge \sum_{i=1}^m \alpha_i y_i = 0, i\in [n],
\end{split}
\end{equation}
where $\gamma = \frac{p}{p-1}>1$, $\theta = C^{1-\gamma} p^{-\gamma} (p-1)\ge 0$. The traditional soft-margin SVM can be viewed as a special case where $\theta = 0$.

\section{Method}

Previous derivations have revealed the generalization bound and dual optimization problem of $p$SVMs. A comparison between Eq. (\ref{eq2}) and Eq. (\ref{dual}) reveals that extending $1$SVM to $p$SVM primarily involves introducing $\theta \sum_{i=1}^m\alpha_i^\gamma$ into the optimization problem. However, this additional term brings barriers to efficient optimization in practical scenarios. 

The Sequential Minimal Optimization (SMO) algorithm \cite{platt1998sequential} is widely recognized for its efficiency in training SVMs by solving large-scale quadratic programming problems. Nevertheless, SMO is specifically designed for the $1$SVM model, i.e. $\theta=0$. In this section, we extend the SMO method to $p$SMO: an SMO-style training algorithm designed for $p$SVMs.

\subsection{$p$SMO Method}

The fundamental principle of SMO is to optimize two variables at a time, keeping the others fixed. Suppose we fix $\alpha_3, \cdots, \alpha_m$ and focus solely on $\alpha_1$ and $\alpha_2$. The new constraints are $\alpha_1,\alpha_2\ge 0\wedge\alpha_1y_1+\alpha_2y_2=c$ where $c$ is a constant. Let $\eta_{12} = K_{11} + K_{12} - 2 K_{12}$ and $E_i=\varphi(\mathbf{x}_i) - y_i$ , where
$$\begin{aligned}
K_{ij}&=K(\mathbf{x}_i, \mathbf{x}_j)=K(\mathbf{x}_j, \mathbf{x}_i)\\
\varphi(\mathbf{x})&=\sum_{j=1}^m \alpha_jy_jK(\mathbf{x}, \mathbf{x}_j) + b\\
v_i&=\sum_{j=3}^m\alpha_j y_jK_{ij}=\varphi(\mathbf{x}_i)-\alpha_1y_1K_{1i} - \alpha_2y_2K_{2i}-b.\\
\end{aligned}$$

The objective function in Eq. (\ref{dual}) can be written as:
\begin{equation}\label{smo1}
\begin{split}&\ \ \ \ \sum_{i=1}^m \alpha_i - \theta\sum_{i=1}^m \alpha_i^\gamma - \dfrac{1}{2} \sum_{i,j=1}^m \alpha_i \alpha_j y_i y_j K(\mathbf{x}_i, \mathbf{x}_j)\\
		&=\alpha_1+\alpha_2 -\theta\alpha_1^\gamma -\theta\alpha_2^\gamma-\dfrac{1}{2}K_{11}\alpha_1^2 - \dfrac{1}{2}K_{22}\alpha_2^2 \\
		&\ \ \ \ \ - \alpha_1\alpha_2y_1 y_2 K_{12}-\alpha_1y_1v_1-\alpha_2y_2v_2 + \text{constant}.\end{split}
\end{equation}

Solving Eq. (\ref{smo1}) under $\alpha_2\ge 0\wedge \alpha_1=cy_1- \alpha_2y_1y_2\ge 0$ yields the update algorithm presented in Algorithm \ref{alg:algorithm}. This algorithm takes $i, j$ as inputs, updating $\alpha_i^{\text{old}}$ and $\alpha_j^{\text{old}}$ to $\alpha_i^{\text{new}}$ and $\alpha_j^{\text{new}}$ in a way that maximizes the objective function while ensuring that $\alpha_i^{\text{old}}y_i+ \alpha_j^\text{old}y_j = \alpha_i^{\text{new}}y_i + \alpha_j^{\text{new}}y_j$ and $\alpha_i^\text{new},\alpha_j^\text{new}\ge 0$. Lemma 2 guarantees that $\eta_{ij}\ge 0$.

\noindent\textbf{Lemma 2} \textit{For any $i, j$, $\eta_{ij}=K_{ii}+K_{jj}-2K_{ij}\ge 0$}.

\begin{algorithm}[tb]
\caption{Update}
\label{alg:algorithm}
\textbf{Input}: Index $i,j$
\begin{algorithmic} 
\STATE Let $c=\alpha_i^{\text{old}}y_i + \alpha_j^{\text{old}}y_j$.
\STATE Let $Q_{ij}=\eta_{ij}\alpha_j^{\text{old}} + y_j(E_i-E_j) - |c|^{\gamma-1}\gamma \theta$.
\STATE Let $g(x)=\eta_{ij}(\alpha_j^{\text{old}} - x) + y_j(E_i-E_j) - x^{\gamma - 1}\gamma \theta + y_iy_j(cy_i-y_iy_jx)^{\gamma-1}\gamma\theta $.
\IF {$y_i=y_j$}
\IF {$-2|c|^{\gamma - 1}\gamma\theta \le Q_{ij}\le \eta_{ij}|c|$}
\STATE $\alpha_j^{\text{new}}\leftarrow$ the unique solution of $g(x)=0$ on $[0, |c|]$.
\ELSE
\STATE $\alpha_j^{\text{new}}\leftarrow\begin{cases}
0, & \text{if } Q_{ij}\le -2|c|^{\gamma-1}\gamma\theta\\
|c|, & \text{if } Q_{ij}\ge \eta_{ij}|c|.
\end{cases}$
\ENDIF
\ELSE
\STATE Let $u=\max(0, -cy_i)$.
\IF {$Q_{ij}\le \eta_{ij} u$}
\STATE $\alpha_j^{\text{new}} \leftarrow u$.
\ELSE
\STATE $\alpha_j^{\text{new}}\leftarrow$ the unique solution of $g(x)=0$ on $[u, +\infty)$.
\ENDIF
\ENDIF
\STATE $\alpha_i^{\text{new}} \leftarrow \alpha_i^{\text{old}} + y_iy_j(\alpha_j^{\text{old}} - \alpha_j^{\text{new}})$.
\STATE Update $(\alpha_i^{\text{old}},\alpha_j^\text{old})$ with $(\alpha_i^{\text{new}}, \alpha_j^{\text{new}})$
\end{algorithmic}
\end{algorithm}

It is evident that the update algorithm coincides with SMO algorithm when $\theta = 0$, suggesting that our $p$SMO method generalizes the SMO method to a broader context.

The core update algorithm has been discussed in Algorithm \ref{alg:algorithm}, and the pseudocode of the $p$SMO algorithm is presented in Listing \ref{algo}. According to the original SMO method, the first index $i$ is selected based on the violation of the KKT conditions, and the second index $j$ is selected to maximize $|E_i - E_j|$. However, in our experiments, we randomly select $j$ because we observed that it is not worthwhile to allocate time to computing $j$ solely for the potential theoretical improvement in convergence speed.

\begin{listing}[tb]%
\caption{$p$SMO algorithm}%
\label{algo}%
\begin{lstlisting}[language=Python]
def fit(X, y, p, C, eps):
  gamma = p / (p - 1)
  theta = ... # Refer to the paper
  for i in range(max_iter):
    Choose i, j to update
    Update(i, j)
    if (the change of alpha < eps):
      break
    Update the value of b
\end{lstlisting}
\end{listing}

\subsection{Updating the Bias Parameter $b$}

In the $p$SMO algorithm, parameter $b$ in Eq. (\ref{eq3}) needs to be updated during each iteration. In this part, we discuss the implementation of this step.

In the original SMO method, the index set $I_{SV}=\{i:0<\alpha_i<C\}$ is used to define the \textit{support vectors} $SV = \{\mathbf{x}_i: i\in I\}$, and $b$ can be computed immediately from any $\mathbf{x}_k\in SV$ as follows:
\begin{equation}\label{updb1}
b = y_k - \sum_{j=1}^m \alpha_j y_j K(\mathbf{x}_j, \mathbf{x}_k).
\end{equation}

However, due to the modification of the original optimization problem, this property no longer holds in the $p$SMO method. Specifically,
\begin{equation}\label{updb2}
    \alpha_k > 0 \not\Rightarrow y_k(\mathbf{w}\cdot \mathbf{x}_k + b) = 1.
\end{equation}

Thus, in our algorithm, an alternative solution is applied:
\begin{equation}\label{updb3}
    b = \dfrac{1}{|I_{SV}|} \sum_{k\in I_{SV}} \Big(y_k - \sum_{j=1}^m \alpha_j y_j K(\mathbf{x}_j, \mathbf{x}_k)\Big),
\end{equation}
where $I_{SV} = \{i:\alpha_i > 0\}$. Or equivalently, $b$ is set to the average bias on all the support vectors.

\subsection{Discussions on Special Cases: $p\in \{1,1.5,2\}$}

While our $p$SMO algorithm is designed for all $p\ge 1$, one of the key challenges in its practical implementation lies in computing the unique solution to $g(x) = 0$. This involves solving the equation:
\begin{equation}\label{eqsolve}
    \alpha_j^{\text{old}} + \dfrac{y_j(E_i-E_j)}{\eta_{ij}} = x - \dfrac{\gamma\theta}{\eta_{ij}}(x^{\gamma - 1} \pm(cy_i\pm x)^{\gamma-1}),
\end{equation}
where $\gamma =\frac{p}{p-1} > 0$. For $p\in\{1,1.5,2\}$, Eq. (\ref{eqsolve}) can be solved analytically. When $p=1$ or $p=2$, Eq. (\ref{eqsolve}) is simplified to a linear equation, and when $p=1.5$, it becomes a quadratic equation. In this cases, implementing our $p$SMO method is more straightforward and retains the same time complexity as the original SMO algorithm. Consequently, in our experiments, we implemented $1.5$SMO and $2$SMO and evaluated their performances.

\subsection{Towards Multiclass Classification}

Previous studies have shown that one-vs-one method may be more suitable than one-vs-rest method for large-scale problems in practical scenarios \cite{hsu2002comparison}. Accordingly, we implemented our multiclass classifier using the one-vs-one strategy, which is based on binary classifiers optimized using $p$SMO.

\section{Experiments}

In this section, we empirically evaluate the effectiveness of our $p$SVM model and $p$SMO method on both binary and multiclass classification tasks. To ensure the reproducibility of our experiments, we set the random seed to 42 for all procedures involving randomness.

\subsection{Binary Classification Tasks}

First, we evaluate the effectiveness of our $p$SVM model on binary classification tasks and explore how the value of $p$ affects the accuracy and the number of support vectors (nSV) on different datasets. 

\subsubsection{Method} To ensure a fair comparison across different values of $p$, we train our $p$SVM model using the cvxpy package \cite{diamond2016cvxpy}, which utilizes the Gurobi optimizer \cite{gurobi} to solve Eq. (\ref{dual}).

\subsubsection{Baselines} Our method is compared with eight classic binary classification algorithms implemented in the scikit-learn package \cite{scikit-learn}, including SVM (RBF SVM, \citealp{boser1992training}), Gaussian Naive Bayes \cite{zhang2004optimality}, Random Forests \cite{breiman2001random}, $k$-Nearest Neighbours \cite{cover1967nearest}, Gradient-boosted Trees \cite{friedman2001greedy}, Linear Discriminant Analysis \cite{fisher1936use}, Decision Trees \cite{Breiman1984ClassificationAR}, and AdaBoost (with Decision Tree estimator, \citealp{hastie2009multi}). 

\subsubsection {Datasets} The datasets selected for evaluation include Cancer (Breast Cancer Wisconsin, \citealp{misc_breast_cancer_wisconsin_(diagnostic)_17}), Heart (Heart Disease, \citealp{misc_heart_disease_45}), Ionosphere (\citealp{misc_ionosphere_52}), Wine (Wine Quality, \citealp{misc_wine_quality_186}), and Banknote (Banknote Authentication, \citealp{misc_banknote_authentication_267}). Their number of features $M$ are reported in Table \ref{table1}. More details of the datasets are provided in the Appendix. 

\subsubsection{Evaluation} Our $p$SVM model and the baseline classifiers are evaluated using the same train-test split. The hyperparameter $C$ in both our model and the baseline SVM is selected via 5-fold cross-validation, with the optimal value $C_{\text{best}}$ reported in Table \ref{table1}. Both our model and the baseline SVM use Gaussian kernel as the kernel function, with its hyperparameter $\sigma$ set to $2\sigma^2 = M\mathrm{Var}[X]$, which is the default setting of the scikit-learn package. Other baseline classifiers are evaluated over 20 runs due to their sensitivity to randomness, with the average accuracy reported in Table \ref{table1}.

\begin{table*}[ht]
\centering{\small
\begin{tblr}{
  cells = {c},
  cell{1}{1} = {r=2}{},
  cell{1}{2} = {r=2}{},
  cell{1}{3} = {r=2}{},
  cell{3}{1} = {r=5}{},
  cell{3}{2} = {r=5}{},
  cell{3}{3} = {r=5}{},
  cell{6}{4} = {r=2}{},
  cell{6}{5} = {c=2}{},
  cell{7}{5} = {c=2}{},
  cell{8}{1} = {r=5}{},
  cell{8}{2} = {r=5}{},
  cell{8}{3} = {r=5}{},
  cell{11}{4} = {r=2}{},
  cell{11}{5} = {c=2}{},
  cell{12}{5} = {c=2}{},
  cell{13}{1} = {r=5}{},
  cell{13}{2} = {r=5}{},
  cell{13}{3} = {r=5}{},
  cell{16}{4} = {r=2}{},
  cell{16}{5} = {c=2}{},
  cell{17}{5} = {c=2}{},
  cell{18}{1} = {r=5}{},
  cell{18}{2} = {r=5}{},
  cell{18}{3} = {r=5}{},
  cell{21}{4} = {r=2}{},
  cell{21}{5} = {c=2}{},
  cell{22}{5} = {c=2}{},
  cell{23}{1} = {r=5}{},
  cell{23}{2} = {r=5}{},
  cell{23}{3} = {r=5}{},
  cell{26}{4} = {r=2}{},
  cell{26}{5} = {c=2}{},
  cell{27}{5} = {c=2}{},
  hline{1,28} = {-}{0.08em},
  hline{3,8,13,18,23} = {-}{0.05em},
}
Setting                & Train         & Test           & $p$                      & 1.00                        & 1.25           & 1.29           & 1.33           & 1.40           & 1.50           & 1.67           & 2.00           & 3.00           \\
                       &               &                & $\gamma = \frac{p}{p-1}$ & /                           & 5              & 4.5            & 4              & 3.5            & 3              & 2.5            & 2              & 1.5            \\
{Cancer\\($M=30$)\\~}  & {398\\(70\%)} & {171\\(30\%)}  & $C_{\text{best}}$        & /                           & 5              & 5              & 5              & 5              & 5              & 5              & 5              & 10             \\
                       &               &                & nSV\%                    & 31.2                        & 31.2           & 30.9           & 31.2           & 31.7           & 31.7           & 32.2           & 34.9           & 39.4           \\
                       &               &                & acc\%                    & \textbf{97.66}              & \textbf{97.66} & \textbf{97.66} & \textbf{97.66} & \textbf{97.66} & \textbf{97.66} & \textbf{97.66} & \textbf{97.66} & \textbf{97.66} \\
                       &               &                & baseline                 & SVM ($C_{\text{best}}=5$)   &                & NB             & RF             & KNN            & GB             & LDA            & DT             & AB             \\
                       &               &                &                          & \textbf{97.66}              &                & 93.57          & 96.93          & 95.91          & 95.91          & 95.32          & 93.13          & 97.08          \\
{Heart\\($M=13$)}      & {189\\(70\%)} & {81\\(30\%)}   & $C_{\text{best}}$        & /                           & 0.5            & 0.5            & 0.5            & 0.5            & 0.5            & 0.5            & 0.5            & 0.1            \\
                       &               &                & nSV\%                    & 78.3                        & 79.4           & 79.4           & 79.4           & 84.1           & 86.2           & 89.4           & 93.7           & 100            \\
                       &               &                & acc\%                    & 82.72                       & \textbf{85.19} & \textbf{85.19} & \textbf{85.19} & \textbf{85.19} & 83.95          & 83.95          & 83.95          & 83.95          \\
                       &               &                & baseline                 & SVM ($C_{\text{best}}=1$)   &                & NB             & RF             & KNN            & GB             & LDA            & DT             & AB             \\
                       &               &                &                          & 82.72                       &                & 83.95          & 80.93          & 80.25          & 78.58          & 83.95          & 72.47          & 81.48          \\
{Ionosphere\\($M=34$)} & {245\\(70\%)} & {106\\(30\%)}  & $C_{\text{best}}$        & /                           & 0.1            & 0.1            & 0.1            & 0.1            & 0.1            & 0.1            & 0.1            & 0.1            \\
                       &               &                & nSV\%                    & 59.2                        & 89.0           & 89.0           & 90.6           & 91.4           & 94.3           & 98.8           & 100            & 100            \\
                       &               &                & acc\%                    & 95.28                       & 96.23          & 96.23          & 96.23          & \textbf{97.17} & \textbf{97.17} & \textbf{97.17} & \textbf{97.17} & \textbf{97.17} \\
                       &               &                & baseline                 & SVM ($C_{\text{best}}=10$)  &                & NB             & RF             & KNN            & GB             & LDA            & DT             & AB             \\
                       &               &                &                          & 96.23                       &                & 88.68          & 95.57          & 83.96          & 94.76          & 84.91          & 88.73          & 92.45          \\
{Wine\\($M=11$)}       & {649\\(10\%)} & {5848\\(90\%)} & $C_{\text{best}}$        & /                           & 0.5            & 0.5            & 0.5            & 0.5            & 0.5            & 0.5            & 0.5            & 0.1            \\
                       &               &                & nSV\%                    & 79.0                        & 79.5           & 79.8           & 79.5           & 82.4           & 84.3           & 87.2           & 91.7           & 100            \\
                       &               &                & acc\%                    & 74.49                       & 74.86          & 74.91          & 74.88          & 75.07          & 75.15          & 75.41          & \textbf{75.63} & 74.91          \\
                       &               &                & baseline                 & SVM ($C_{\text{best}}=5$)   &                & NB             & RF             & KNN            & GB             & LDA            & DT             & AB             \\
                       &               &                &                          & 74.56                       &                & 70.09          & 75.50          & 69.82          & 75.48          & 73.58          & 67.42          & 73.12          \\
{Banknote\\($M=4$)}    & {411\\(30\%)} & {961\\(70\%)}  & $C_{\text{best}}$        & /                           & 0.5            & 0.5            & 0.5            & 0.5            & 0.5            & 0.5            & 0.5            & 1              \\
                       &               &                & nSV\%                    & 26.3                        & 40.6           & 41.6           & 45.5           & 48.4           & 49.1           & 56.2           & 66.4           & 82.2           \\
                       &               &                & acc\%                    & \textbf{100.0}              & \textbf{100.0} & \textbf{100.0} & \textbf{100.0} & \textbf{100.0} & \textbf{100.0} & \textbf{100.0} & \textbf{100.0} & \textbf{100.0} \\
                       &               &                & baseline                 & SVM ($C_{\text{best}}=0.5$) &                & NB             & RF             & KNN            & GB             & LDA            & DT             & AB             \\
                       &               &                &                          & 99.38                       &                & 85.02          & 98.47          & 98.86          & 98.96          & 97.29          & 96.93          & 98.96          
\end{tblr}}
\caption{Evaluation results on binary classification tasks.}
\label{table1}
\end{table*}

\begin{figure*}[ht]
\centering
\includegraphics[width=0.95\textwidth]{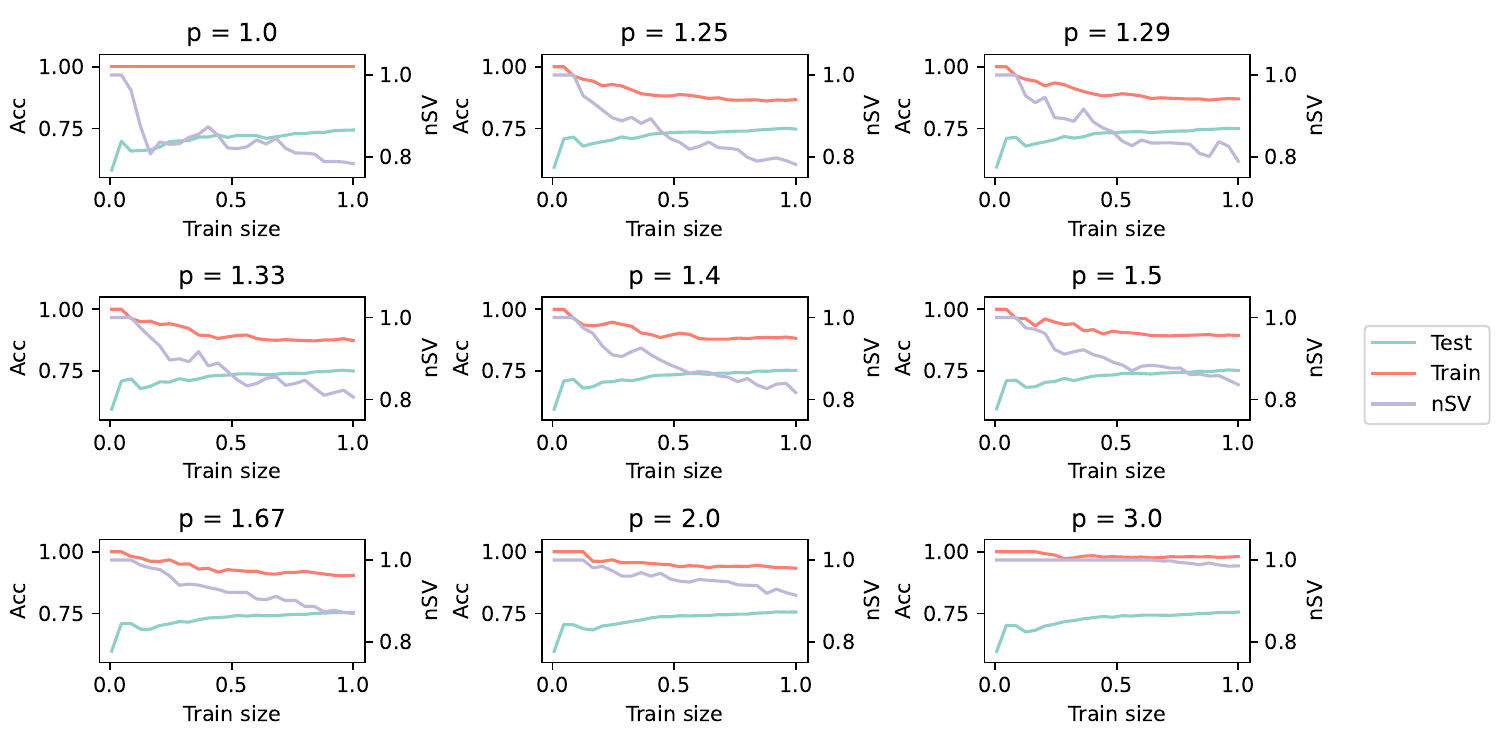} 
\caption{Training process on the Wine Quality Dataset.}
\label{fig2}
\end{figure*}

\subsubsection{Results} The evaluation results are reported in Table \ref{table1}, with the highest performance on each dataset highlighted in bold. The training process on the Wine Dataset is depicted in Fig. \ref{fig2}, where ``Train size" denotes the proportion of the training sample used for training. Our experimental findings lead to several key conclusions:

\begin{itemize}
    \item Compared to other widely used classification algorithms, $p$SVM achieves the best classification performance on all selected datasets. This is largely due to the introduction of the free parameter $p$, which brings more flexibility to the traditional SVM. Notably, the best accuracy of our $p$SVM model on the Heart Disease Dataset beats MeanMap \cite{quadrianto2008estimating}, InvCal \cite{inproceedings}, conv-$\propto$SVM and alter-$\propto$SVM \cite{yu2013svm}. However, introducing the term $\theta \sum_{i=1}^m \alpha_i^{\gamma}$ increases the training time, especially when $p\not\in \{1, 1.5, 2\}$. Therefore, introducing the $p$SMO method is essential towards making large-scale training more feasible.
    \item When $p=1$, the optimization problem (Eq. \ref{dual}) has no upper-bound on $\alpha_i$, which can be interpreted as a traditional soft-margin SVM where $C=+\infty$. Therefore, our $1$SVM is actually a ``hard" SVM which has $100\%$ accuracy on the training sample which can be linearly separated (using Gaussian Kernel) as reported in Fig. \ref{fig2}, leading to its lower accuracy than traditional soft-margin SVM as reported in Table \ref{table1}. 
    \item As $p$ increases, nSV increases accordingly, suggesting that the model learns better on the training set, but at the cost of an increased risk of overfitting. The results in Table \ref{table1} support this conclusion: larger $p$ values lead to higher nSV, however, the test accuracy initially increases and then decreases, indicating that the model initially learns the features of the training set well, but begins to overfit as $p$ continues to grow. Based on our experiments, we infer that our $p$SVM model performs optimally when $p\in[1.25, 2]$. Therefore, one can determine the best $p$ value within this range using cross-validation, or simply select $p=1.5$ or $p=2$ and apply $1.5$SMO or $2$SMO for faster training.
\end{itemize}

\subsection{Multiclass Classification Tasks}

Next, we evaluate the effectiveness of our $p$SVM model and $p$SMO method on multiclass classification tasks.

\subsubsection{Preliminaries} Recent advances in multiclass classification have led to the development of powerful methods, such as CappedSVM \cite{nie2017multiclass} and the state-of-the-art method $\text{M}^3$SVM \cite{nie2024multi}. To facilitate a comparative analysis, we draw on the data from \citealp{nie2024multi} to benchmark our model against these established approaches.

\subsubsection{Baselines}

Our model was evaluated against the same set of baselines as used in \citealp{nie2024multi}, which includes eight multiclass classification algorithms, including one-vs-rest (or one-vs-all, OvR), one-vs-one (OvO), Crammer \cite{crammer2001algorithmic}, M-SVM \cite{bredensteiner1999multicategory}, Top-k \cite{lapin2015top}, Multi-LR \cite{bohning1992multinomial}, Sparse Multinomial Logistic Regression (SMLR, \citealp{krishnapuram2005sparse}), and $\text{M}^3$SVM \cite{nie2024multi}.

\begin{table}[ht]
\centering{\small
\begin{tblr}{
  cells = {c},
  hline{1,6} = {-}{0.08em},
  hline{2} = {-}{0.05em},
}
Dataset     & Instances & Features & Classes \\
Glass       &   214    &     9     &     6    \\
Vehicle     &    845       &   18       &     4    \\
Dermatology &     358      &    34      &     6    \\
USPS        &    9298       &    256      &    10     
\end{tblr}
}
\caption{Details of multiclass classification datasets.}
\label{table3}
\end{table}

\subsubsection{Datasets} The datasets chosen for evaluation include Glass \cite{misc_glass_identification_42}, Vehicle \cite{misc_statlog_(vehicle_silhouettes)_149}, Dermatology \cite{misc_dermatology_33}, and USPS \cite{291440}. The properties of these datasets are summarized in Table \ref{table3}. Additional details about these datasets are provided in the Appendix.

\begin{table*}[ht]
\centering{\small
\begin{tblr}{
  cells = {c},
  hline{1,6} = {-}{0.08em},
  hline{2} = {-}{0.05em},
}
Train & Test & Methods     & OvR   & OvO   & Crammer & M-SVM & Top-k & Multi-LR & SMLR  & $\text{M}^3\text{SVM}$ & 1.5L           & 2L             \\
171   & 43   & Glass       & 0.656 & 0.685 & 0.594   & 0.629 & 0.674 & 0.664    & 0.679 & 0.744                  & 0.744          & \textbf{0.767} \\
676   & 169  & Vehicle     & 0.794 & 0.756 & 0.757   & 0.762 & 0.778 & 0.780    & 0.771 & 0.800         & \textbf{0.834}          & 0.828          \\
286   & 72   & Dermatology & 0.939 & 0.971 & 0.933   & 0.868 & 0.891 & 0.965    & 0.965 & \textbf{0.986}         & \textbf{0.986} & \textbf{0.986} \\
7438  & 1860 & USPS        & 0.887 & 0.898 & 0.769   & 0.910 & 0.825 & 0.932    & 0.937 & 0.956                  & 0.959          & \textbf{0.960} 
\end{tblr}
}
\caption{Evaluation results on multiclass classification tasks.}
\label{table2}
\end{table*}
\subsubsection{Evaluation}

We follow the same experimental settings outlined in \citealp{nie2024multi}. Both our $p$SMO algorithm and the baseline algorithms are evaluated using the same 8:2 train-test split, with the test accuracy reported in Table \ref{table2}. All the algorithms are trained using a linear kernel. In Table \ref{table2}, ``1.5L'' stands for ``$1.5$SMO + Linear Kernel'', and ``2L'' stands for ``$2$SMO + Linear Kernel''. We report the performance of $p$SMO on $C_{\text{best}}$ selected from grid search.

\subsubsection{Results} The evaluation results are presented in Table \ref{table2}, with the highest performance on each dataset highlighted in bold. Compared to other multiclass classification algorithms, our $p$SMO algorithm demonstrates superior performance, particularly on the USPS Dataset.

Our experimental findings lead to two key conclusions: Firstly, the OvO method shows improved performance when the base classifier is replaced with our $p$SVM model. Secondly, our $p$SMO algorithm surpasses the state-of-the-art method across various datasets.

\section{Conclusions}

Soft-margin SVMs are well-established in various classification tasks and algorithms, yet the optimal norm $p$ for hinge loss has received limited attention. To address this, we introduced the $p$SVM model and the $p$SMO algorithm, designed to deliver both precise classification and efficient training. Through theoretical analysis and empirical experimentation, we derived the generalization bounds for $p$SVMs and detailed the practical implementations of $1.5$SMO and $2$SMO. Our experimental results on multiple datasets highlight the superior performance of our methods compared to traditional and state-of-the-art approaches.

\bibliography{aaai25}

\appendix

\section{Appendix}

\subsection{A.1 Proof}

\noindent\textbf{Lemma 1} \textit{$\Phi_{p,\rho}$ is $p/\rho$-lipschitz}.

\noindent\textit{Proof.} Assuming $x_1<x_2$.

\begin{itemize}
    \item $x_1<0, x_2\in [0, \rho]$, we have
    $$\begin{aligned}|\Phi(x_1) - \Phi(x_2)| &= 1- \Big(1-\frac{x_2}{\rho}\Big)^p\le \dfrac{p}{\rho}x_2 \\ &\le \dfrac{p}{\rho} |x_1 - x_2|.\end{aligned}$$
    \item $x_1 <0, x_2\ge \rho$, we have
    $$|\Phi(x_1)-\Phi(x_2)|=1\le \dfrac{p}{\rho} |x_1 - x_2|.$$
    \item $x_1\in[0,\rho], x_2\in [0, \rho]$, we have
    $$\begin{aligned}|\Phi(x_1)-\Phi(x_2)|&=\Big(1-\frac{x_1}{\rho}\Big)^p- \Big(1-\frac{x_2}{\rho}\Big)^p\\
        &\le \dfrac{p}{\rho} |x_1-x_2|,\end{aligned}$$
    using the fact that $(1-\frac{x}{\rho})^p + \frac{p}{\rho}x$ is monotonically increasing on $[0,\rho]$.
    \item $x_1\in[0, \rho], x_2 \ge \rho$, we have
    $$\begin{aligned}|\Phi(x_1) - \Phi(x_2)| &= \Big(1-\frac{x_1}{\rho}\Big)^p\le \dfrac{p}{\rho}(\rho - x_1) \\ &\le \dfrac{p}{\rho} |x_1 - x_2|,\end{aligned}$$
    using the fact that $(1-\frac{x}{\rho})^p + \frac{p}{\rho}x$ is monotonically increasing on $[0,\rho]$.
\end{itemize}
\hfill{$\square$}

\noindent\textbf{Claim 1} \textit{Let $S\subseteq\{\mathbf{x}: \Vert\mathbf{x}\Vert\le r\}$ be a sample of size $m$ and let $\mathcal{H}=\{\mathbf{x}\mapsto \mathbf{w}\cdot \mathbf{x} + b: \Vert\mathbf{w}\Vert \le \Lambda\}$. Then, the empirical Rademacher complexity of $\mathcal{H}$ can be bounded as follows}:
$$\widehat{\mathfrak{R}}_{S}(\mathcal{H})\le \sqrt{\dfrac{r^2\Lambda^2}{m}}.$$

\textit{Proof}.
$$\begin{aligned}\widehat{\mathfrak{R}}_{S}(\mathcal{H}) &= \dfrac{1}{m} \underset{\sigma}{\mathbb{E}}\Big[\underset{\Vert{\mathbf{w}}\Vert\le \Lambda}{\sup}\sum_{i=1}^m \sigma_i(\mathbf{w}\cdot\mathbf{x}_i + b)\Big]\\ &=\dfrac{1}{m} \underset{\sigma}{\mathbb{E}}\Big[\underset{\Vert{\mathbf{w}}\Vert\le \Lambda}{\sup} \mathbf{w}\cdot\sum_{i=1}^m \sigma_i\mathbf{x}_i\Big]\\&\le \dfrac{\Lambda}{m} \underset{\sigma}{\mathbb{E}}\Big[\Big\Vert\sum_{i=1}^m \sigma_i \mathbf{x}_i\Big\Vert\Big]\le \dfrac{\Lambda}{m} \Big[\underset{\sigma}{\mathbb{E}}\Big[\Big\Vert\sum_{i=1}^m \sigma_i \mathbf{x}_i\Big\Vert^2\Big]\Big]^{\frac{1}{2}}\\&=\dfrac{\Lambda}{m} \Big[\underset{\sigma}{\mathbb{E}}\Big[\sum_{i,j=1}^m \sigma_i \sigma_j (\mathbf{x}_i\cdot\mathbf{x}_j)\Big]\Big]^{\frac{1}{2}}\\&\le \dfrac{\Lambda}{m} \Big[\sum_{i=1}^m \Vert{\mathbf{x}_i}\Vert^2\Big]^{\frac{1}{2}}\le \sqrt{\dfrac{r^2\Lambda^2}{m}}.\end{aligned}$$\hfill{$\square$}

\noindent\textbf{Claim 2} \textit{Let $S\subseteq\{\mathbf{x}: \Vert\mathbf{x}\Vert\le r\}$ be a sample of size $m$ and let $\mathcal{H}$ be a set of real-valued functions. Fix $\rho >0$, then, for any $\delta >0$, with probability at least $1-\delta$, the following holds for any $h\in \mathcal{H}$}:
$$R_{p,\rho}(h)\le \widehat{R}_{S,p,\rho}(h) + \dfrac{2p}{\rho} \mathfrak{R}_m(\mathcal{H})+\sqrt{\dfrac{\log \frac{1}{\delta}}{2m}}.$$

\textit{Proof}. Use the same proof method as in Theorem 5.8 of \citealp{mohri2018foundations}, and the only difference is that $\Phi=\Phi_{p,\rho}$ is $p/\rho$-lipschitz. \hfill{$\square$}

\noindent\textbf{Claim 3} \textit{Let $S\subseteq\{\mathbf{x}: \Vert\mathbf{x}\Vert\le r\}$ be a sample of size $m$ and let $\mathcal{H}$ be a set of real-valued functions. Fix $r >0$, then, for any $\delta >0$, with probability at least $1-\delta$, the following holds for any $h\in \mathcal{H}$ and $\rho\le r$}:
$$\begin{aligned}R_{p,\rho}(h)\le &\ \widehat{R}_{S,p,\rho}(h)\\ &+ \dfrac{4p}{\rho}\mathfrak{R}_m(\mathcal{H})+\sqrt{\dfrac{\log \log_2 \frac{2r}{\rho}}{m}} + \sqrt{\dfrac{\log \frac{2}{\delta}}{2m}}.\end{aligned}$$

\textit{Proof}. Use the same proof method as in Theorem 5.9 of \citealp{mohri2018foundations}, and the only difference is that $\Phi=\Phi_{p,\rho}$ is $p/\rho$-lipschitz. \hfill{$\square$}

\noindent\textbf{Theorem 3} \textit{Let $\mathcal{H} = \{\mathbf{x}\mapsto \mathbf{w}\cdot\mathbf{x} + b: \Vert\mathbf{w}\Vert\le 1\}$ and assume that $\mathcal{X} \subseteq \{\mathbf{x}:\Vert\mathbf{x}\Vert\le r\}$. Fix $r$, then, for any $\delta > 0$, with probability at least $1-\delta$ over the choice of a sample $S$ of size $m$, the following holds for any $h\in \mathcal{H}$ and $\rho\in (0,r]$}:
$$\begin{aligned}R_{p,\rho}(h)\le &\ \widehat{R}_{S,p,\rho}(h)\\ &+ 4p\sqrt{\dfrac{r^2/\rho^2}{m}}+\sqrt{\dfrac{\log \log_2 \frac{2r}{\rho}}{m}} + \sqrt{\dfrac{\log \frac{2}{\delta}}{2m}}.\end{aligned}$$

\textit{Proof}. Let $\Lambda = 1$ in Claim 1 and use it in Claim 3 directly to get the upper-bound. \hfill{$\square$}

\noindent\textbf{Corollary 1} (Generalization bound of $p$SVMs) \textit{Suppose $\mathcal{H} = \{\mathbf{x}\mapsto \mathbf{w}\cdot\mathbf{x} + b: \Vert\mathbf{w}\Vert\le 1/\rho\}$ and assume that $\mathcal{X} \subseteq \{\mathbf{x}:\Vert\mathbf{x}\Vert\le r\}$. Fix $r>0$, then, for any $\delta > 0$, with probability at least $1-\delta$ over the choice of a sample $S$ of size $m$, the following holds for any $h\in \mathcal{H}$ and $\rho\in(0,r]$}:
$$\begin{aligned}R_{p,\rho}(h)\le &\ \dfrac{1}{m}\sum_{i=1}^m (\max(0, 1-y_i(\mathbf{w}\cdot \mathbf{x}_i + b)))^p\\ &+ 4p\sqrt{\dfrac{r^2/\rho^2}{m}}+\sqrt{\dfrac{\log \log_2 \frac{2r}{\rho}}{m}} + \sqrt{\dfrac{\log \frac{2}{\delta}}{2m}}.\end{aligned}$$

\textit{Proof}. $\Phi_{p,\rho}$ is upper-bounded by the $p$-norm $\rho$-hinge loss:
$$\Phi_{p,\rho}(x)\le \max(0, 1-x/\rho)^p,$$

Thus, let $\mathcal{H} = \{\mathbf{x}\mapsto \mathbf{w}\cdot\mathbf{x} + b: \Vert\mathbf{w}\Vert\le 1\}$ and assume that $\mathcal{X} \subseteq \{\mathbf{x}:\Vert\mathbf{x}\Vert\le r\}$. Fix $r>0$, then, for any $\delta > 0$, with probability at least $1-\delta$ over the choice of a sample $S$ of size $m$, the following holds for any $h\in \mathcal{H}$ and $\rho\in(0,r]$:
$$\begin{aligned}R_{p,\rho}(h)\le &\ \dfrac{1}{m}\sum_{i=1}^m (\max(0, 1-\frac{y_i(\mathbf{w}\cdot \mathbf{x}_i + b)}{\rho}))^p\\ &+ 4p\sqrt{\dfrac{r^2/\rho^2}{m}}+\sqrt{\dfrac{\log \log_2 \frac{2r}{\rho}}{m}} + \sqrt{\dfrac{\log \frac{2}{\delta}}{2m}}.\end{aligned}$$

Here we need $\Vert\mathbf{w}\Vert\le1$. However, if we consider $\mathbf{w}'=\mathbf{w}/\rho$ and $b'=b/\rho$, we can directly derive Corollary 1.\hfill{$\square$}

\subsubsection{Dual Problem Derivation} The dual optimization problem of our $p$SVM model is as follows.
$$\begin{aligned}\max_{\boldsymbol{\alpha}}\ \ \ &\sum_{i=1}^m \alpha_i - \theta\sum_{i=1}^m \alpha_i^\gamma - \dfrac{1}{2} \sum_{i,j=1}^m \alpha_i \alpha_j y_i y_j K(\mathrm{x}_i, \mathrm{x}_j), \\ \text{s.t.} &\ \alpha_i \ge 0 \wedge \sum_{i=1}^m \alpha_i y_i = 0, i\in [n],\end{aligned}$$
where $\gamma = \frac{p}{p-1}>1$, $\theta = C^{1-\gamma} p^{-\gamma} (p-1)\ge 0$.

\noindent\textit{Proof}. We introduce Lagrange variables $\alpha_i ,\beta_i\ge 0, i\in [m]$, so the Lagrangian can be defined by
	$$\begin{aligned}\mathcal{L}(\mathbf{w},b,\boldsymbol{\xi},\boldsymbol{\alpha},\boldsymbol{\beta})=&\ \ \frac12\|\mathbf{w}\|^2+C\sum_{i=1}^m\xi_i^p-\sum_{i=1}^m\beta_i\xi_i\\
		&-\sum_{i=1}^m\alpha_i[y_i(\mathbf{w}\cdot\mathbf{x}_i+b)-1+\xi_i].\end{aligned}$$
	
	The KKT conditions are as follows.
	
	$$\begin{aligned}
		&\nabla_{\xi_i} \mathcal{L} = Cp\xi_{i}^{p - 1} - \beta_i - \alpha_i = 0\\
		&\nabla_{\mathbf{w}} \mathcal{L} = \mathbf{w} - \sum_{i=1}^m \alpha_i y_i \mathbf{x}_i = 0\\
		&\nabla_b \mathcal{L} = -\sum_{i=1}^m \alpha _i y_i = 0\\
		&\beta_i = 0 \lor \xi_i = 0, \forall i\in [m]\\
		&\alpha_i = 0 \lor y_i(\mathbf{w}\cdot\mathbf{x}_i + b) = 1 - \xi_i, \forall i\in [m]
	\end{aligned}$$
	
	By plugging the equations into the Lagrangian, we have
	$$\mathcal{L} = 
	\sum_{i=1}^m \alpha_i - \dfrac{1}{2}\Vert\sum_{i=1}^m \alpha_iy_i\mathbf{x}_i\Vert^2 - C(p-1)\sum_{i=1}^m\xi_i^p.$$
	
	When we choose the traditional $1$-norm penalty in SVMs, we have
	$$\mathcal{L} = 
	\sum_{i=1}^m \alpha_i - \dfrac{1}{2}\Vert\sum_{i=1}^m \alpha_iy_i\mathbf{x}_i\Vert^2,$$ so our optimization problem is $$\begin{aligned}\max_{\boldsymbol{\alpha}}\ \ \ &\sum_{i=1}^m \alpha_i - \dfrac{1}{2}\Vert\sum_{i=1}^m \alpha_iy_i\mathbf{x}_i\Vert^2\\ \mathrm{subject\ to}\ \ \ &\alpha_i \in [0, C] \wedge \sum_{i=1}^m \alpha_i y_i = 0,i\in[m],\end{aligned},$$
	which has been widely discussed in previous works. 
	
	Now we consider the SVMs with $p$-norm loss and $p>1$. Consider the set of indexes $I = \{i: \xi_i = 0\}$, we can infer that $\forall i \in I, \alpha_i = \beta_i = 0$, and $\forall i \not\in I, \alpha_i = Cp\xi_{i}^{p-1}$, leading to a common conclusion that
	$$\alpha_i = Cp\xi_{i}^{p-1}, \forall i\in [m].$$
	
	So our optimization problem becomes
	$$\begin{aligned}\max_{\boldsymbol{\alpha}}\ \ \ &\sum_{i=1}^m \alpha_i - \dfrac{1}{2}\Vert\sum_{i=1}^m \alpha_iy_i\mathbf{x}_i\Vert^2 \\
		&- C(p - 
		1)\sum_{i=1}^m\Big(\dfrac{\alpha_i}{Cp}\Big)^{\frac{p}{p-1}}\\ \mathrm{subject\ to}\ \ \ &\alpha_i \ge 0 \wedge \sum_{i=1}^m \alpha_i y_i = 0,i\in[m].\end{aligned}$$
	
	We can rewrite our objective function as follows, assuming that $\gamma = \frac{p}{p-1}>1,\theta = C^{1-\gamma}p^{-\gamma}(p-1)\ge0$, then
	$$G=\sum_{i=1}^m \alpha_i - \dfrac{1}{2} \Vert\sum_{i=1}^m \alpha_i y_i \mathbf{x}_i\Vert^2 - \theta\sum_{i=1}^m \alpha_i^\gamma.$$

\subsubsection{$p$SMO Algorithm Derivation}

The key idea of SMO is: freeze all the variables except two of them, and look at only two variables. Assume that we are freezing $\alpha_3, \cdots, \alpha_m$ and we only consider $\alpha_1$ and $\alpha_2$. New constraints are $\alpha_1,\alpha_2\ge 0\wedge\alpha_1y_1+\alpha_2y_2=c$ where $c$ is a constant. Let $\eta_{12} = K_{11} + K_{12} - 2 K_{12}$ and $E_i=\varphi(\mathbf{x}_i) - y_i$ , where
$$\begin{aligned}
K_{ij}&=K(\mathbf{x}_i, \mathbf{x}_j)=K(\mathbf{x}_j, \mathbf{x}_i)\\
\varphi(\mathbf{x})&=\sum_{j=1}^m \alpha_jy_jK(\mathbf{x}, \mathbf{x}_j) + b\\
v_i&=\sum_{j=3}^m\alpha_j y_jK_{ij}=\varphi(\mathbf{x}_i)-\alpha_1y_1K_{1i} - \alpha_2y_2K_{2i}-b.\\
\end{aligned}$$
	
	The optimization problem can be written as follows:
	$$\begin{aligned}G&=\sum_{i=1}^m \alpha_i - \theta\sum_{i=1}^m \alpha_i^\gamma - \dfrac{1}{2} \sum_{i,j=1}^m \alpha_i \alpha_j y_i y_j K(\mathrm{x}_i, \mathrm{x}_j)\\
		&=\alpha_1+\alpha_2 -\theta\alpha_1^\gamma -\theta\alpha_2^\gamma-\dfrac{1}{2}K_{11}\alpha_1^2 - \dfrac{1}{2}K_{22}\alpha_2^2 \\
		&\ \ \ \ \ - \alpha_1\alpha_2y_1 y_2 K_{12}-\alpha_1y_1v_1-\alpha_2y_2v_2 + \mathrm{constant}.\end{aligned}$$

\noindent\textbf{Lemma 2} \textit{For any $i, j$, $\eta_{ij}=K_{ii}+K_{jj}-2K_{ij}\ge 0$}.

\noindent\textit{Proof.} Using the Cauchy-Schwarz inequality on the given reproducing kernel Hilbert space (RKHS), we have
$$2K(x,y)\le 2\sqrt{K(x,x)K(y,y)}\le K(x,x)+K(y,y).\ \square$$

We want to update $\alpha_1^{\mathrm{old}}$ and $\alpha_2^{\mathrm{old}}$ into $\alpha_1^{*}$ and $\alpha_2^{*}$. It is obvious that
 $$\alpha_1^{\mathrm{old}}y_1 + \alpha_2^{\mathrm{old}}y_2 = \alpha_1^{*}y_1 + \alpha_2^{*}y_2=c,$$
 or equivalently, 
 $$\alpha_1^{\mathrm{old}} + y_1y_2\alpha_2^{\mathrm{old}} = \alpha_1^{*} + y_1y_2\alpha_2^{*}=cy_1.$$

Plugging the equation above into $G$ transfers $G$ into a single variable optimization problem, i.e. $G=W(\alpha_2^{*})$ as follows.
 $$\begin{aligned}W'(\alpha_2^{*})=&\ \eta_{12}(\alpha_2^{\mathrm{old}} - \alpha_2^{*}) + y_2(E_1 - E_2)\\ &+ y_1 y_2 (cy_1 - y_1y_2\alpha_2^{*})^{\gamma - 1}\gamma\theta - (\alpha_2^{*})^{\gamma - 1} \gamma \theta.\end{aligned}$$

 So we have
 $$\begin{aligned}W''(\alpha_2^*) =&\ -\eta_{12} - \gamma\theta(\gamma - 1)(cy_1 - y_1y_2 \alpha_2^*)^{\gamma - 2} \\ &\ - \gamma\theta(\gamma - 1)(\alpha_2^*) ^{\gamma - 2}.\end{aligned}$$

Let $D=\{\alpha_2: \alpha_2\ge 0 \ \wedge\ \alpha_1 = cy_1 - y_1y_2 \alpha_2\ge 0\}$. For any $\alpha_2^* \in D$, we have $W''(\alpha_2^*) \le 0$. 

Let $g(x) = W'(x)$, then $g$ is monotonically decreasing on $D$. We want to find out $\alpha_1^{\mathrm{new}}$ and $\alpha_2^{\mathrm{new}}$ which maximize $G(\alpha_1^*, \alpha_2^*) = W(\alpha_2^*)$. We consider the following cases and discuss them separatedly.

\textbf{Case 1.} $y_1 = y_2$

In this case, we have $D=[0, cy_1]$. We can immediately derive that
$$g(0) = \eta_{12} \alpha_2^{\mathrm{old}} + y_2(E_1 - E_2) + |c|^{\gamma - 1}\gamma \theta,$$
$$g(cy_1) = \eta_{12}(\alpha_2^{\mathrm{old}} - cy_1) + y_2(E_1-E_2) - |c|^{\gamma - 1} \gamma\theta.$$

\begin{itemize}

\item $g(0) \le 0$. $\alpha_2^{\mathrm{new}}= 0$.

\item $g(cy_1) \ge 0$. $\alpha_2^{\mathrm{new}} = cy_1$. 

\item $g(0) > 0\ \wedge\ g(cy_1) < 0$. $\alpha_2^{\mathrm{new}}$ satisfies $g(\alpha_2^{\mathrm{new}})=0$.

\end{itemize}

\textbf{Case 2.} $y_1 \neq y_2$

In this case, let $a=\max\{0, -cy_1\}$, we have $D=[a, +\infty)$. We can immediately derive that
$$g(0) = \eta_{12}\alpha_2^{\mathrm{old}} + y_2(E_1 - E_2) - |c|^{\gamma - 1}\gamma \theta,$$
$$g(-cy_1)= \eta_{12}(\alpha_2^{\mathrm{old}}+ cy_1) + y_2(E_1 - E_2) - |c|^{\gamma - 1}\gamma\theta.$$

\begin{itemize}
\item $g(a)\le 0$. $\alpha_2^{\mathrm{new}} = a$.

\item$g(a)> 0$. $\alpha_2^{\mathrm{new}}$ satisfies $g(\alpha_2^{\mathrm{new}}) = 0$.

\end{itemize}

\textbf{Generalized version}

We can summarize our case analysis above as follows. Let
$$Q_{12}=\eta_{12}\alpha_2^{\mathrm{old}} + y_2(E_1 - E_2) - |c|^{\gamma - 1}\gamma \theta.$$

1. $y_1 = y_2$.

\begin{itemize}

\item $Q_{12}\le -2|c|^{\gamma - 1}\gamma \theta$. $\alpha_2^{\mathrm{new}}=0$.

\item $Q_{12}\ge \eta_{12}|c|$. $\alpha_2^{\mathrm{new}} = |c|$.

\item Otherwise, solve $g(x)=0$ on $[0, |c|]$ to get $\alpha_2^{\mathrm{new}}$.

\end{itemize}

2. $y_1\neq y_2$. Let $a=\max\{0, -cy_1\}$.

\begin{itemize}

\item $Q_{12}\le \eta_{12}a$. $\alpha_2^{\mathrm{new}}=a$.

\item Otherwise, solve $g(x)=0$ on $[a, +\infty)$ to get $\alpha_2^{\mathrm{new}}$.

\end{itemize}

\subsection{A.2 More on Experiments}

\subsubsection{Training process on USPS Dataset}

Figure 3 illustrates the training process on USPS with $p=2$ and $C=0.25$.

\subsubsection{Datasets}

All the datasets used in our experiments were cleaned and scaled. We removed any rows with missing values and applied the sklearn.preprocessing.scale function to standardize the data.

For binary classification datasets, labels were standardized to $\{-1,1\}$. In the case of the Wine Quality dataset, where labels represent a quality level from $[0,10]$, we categorized samples with quality levels in the range $[0,5]$ as negative and those in $[6,10]$ as positive.

\subsubsection{Reproducibility}

All the results presented in this paper can be reproduced using the provided code. To run our code, please ensure that all dependencies listed in requirements.txt are installed. Please note the following:

\begin{itemize}

\item The cupy package is utilized to accelerate matrix computations on GPUs. If you prefer to use numpy, please refer to the code for instructions on how to adapt it accordingly.

\item A Gurobi license (e.g., Academic License) is required to run our code, as the Gurobi optimizer is used in the experiments.

\end{itemize}

To reproduce our results, please run ``python eval.py'' under ``binary'' and ``multiclass'' folder in the code we provided.

\begin{figure}[t]
\centering
\includegraphics[width=0.9\columnwidth]{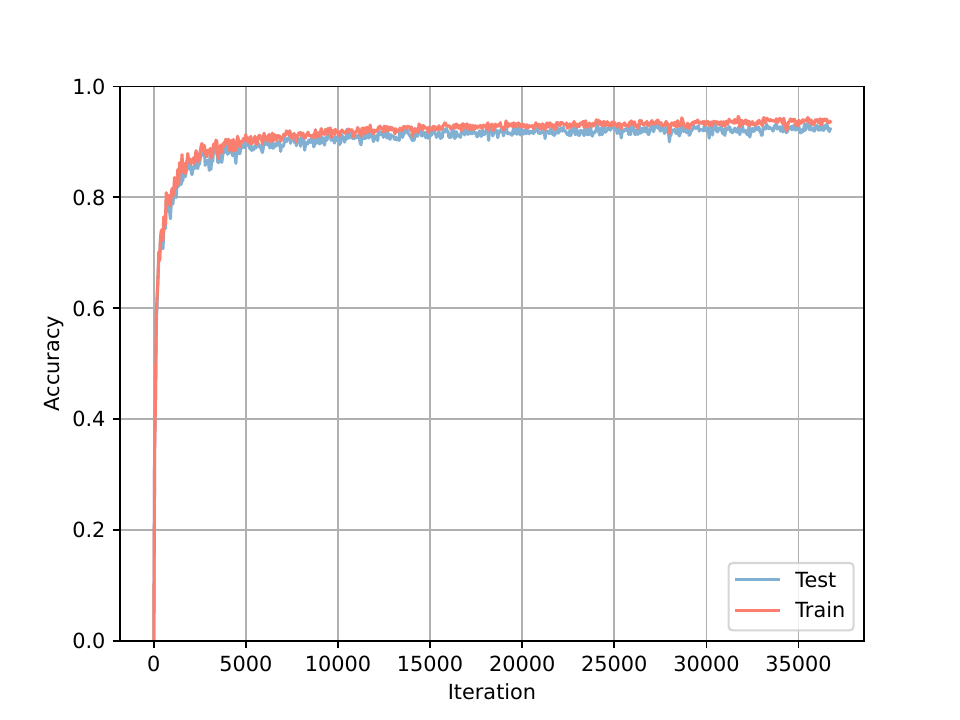} 
\caption{Training process on USPS with $p=2,C=0.25$.}
\label{figtrain}
\end{figure}

\end{document}